%% file: main.tex
\documentclass[runningheads]{llncs}

\usepackage{eccv}

\usepackage{eccvabbrv}

\usepackage{graphicx}
\usepackage{booktabs}

\usepackage[accsupp]{axessibility}  

\usepackage[pagebackref,breaklinks,colorlinks,citecolor=eccvblue]{hyperref}

\usepackage{orcidlink}
\usepackage{multirow}
\usepackage{pifont}
\usepackage{xcolor}
\usepackage{colortbl}
\usepackage{url}
\usepackage{bbm}
\usepackage{amsmath}        
\usepackage{mathrsfs}
\usepackage{physics}
\usepackage{amsfonts}       
\usepackage{wrapfig} 

\newcommand\blfootnote[1]{%
  \begingroup
  \renewcommand\thefootnote{}\footnote{#1}%
  \addtocounter{footnote}{-1}%
  \endgroup
}

\input{Texs/notations}

\title{{\Dataset}: A Benchmark and Model for Universal 6D Object Pose Estimation and Tracking} 

\titlerunning{Omni6DPose}

\author{Jiyao Zhang\inst{1, 2, 3*} \and
Weiyao Huang\inst{1*} \and
Bo Peng\inst{1*} \and
Mingdong Wu\inst{1, 2, 3} \and
Fei Hu\inst{1, 3} \and
Zijian Chen\inst{4} \and
Bo Zhao\inst{5} \and
Hao Dong\inst{1, 2, 3\dagger}}

\authorrunning{J. Zhang, W. Huang, B. Peng, M. Wu, F. Hu, Z. Chen, B. Zhao, H. Dong}

\institute{\scriptsize
Center on Frontiers of Computing Studies, School of Computer Science, Peking University \and
PKU-Agibot Lab, School of Computer Science, Peking University \and
National Key Laboratory for Multimedia Information Processing, School of Computer Science, \\
\mbox{Peking University \and Waseda University \and Beijing Academy of Artificial Intelligence} \\
\email{\{jiyaozhang, sshwy, bo.peng, hofee\}@stu.pku.edu.cn} \\
\email{zijianchenwaseda@akane.waseda.jp}
\email{\{wmingd, bozhao, hao.dong\}@pku.edu.cn}
}

\begin{document}
\maketitle

\begin{abstract}
    \vspace{-20pt}
    \input{Texs/0_abstract}
    \vspace{-8pt}
    \blfootnote{*: equal contributions, $\dagger$: corresponding author}
\end{abstract}


\vspace{-32pt}
\section{Introduction}
\vspace{-10pt}
\label{sec:intro}
\input{Texs/1_intro}

\vspace{-10pt}
\section{Related work}
\label{sec:related}
\input{Texs/2_related_work}

\vspace{-10pt}
\section{{\Dataset} Dataset}
\vspace{-10pt}
\input{Texs/3_data_creation}

\vspace{-10pt}
\section{Category-level 6D Pose Estimation Method}
\input{Texs/4_method}

\vspace{-10pt}
\section{Experiments}
\input{Texs/5_experiments}

\vspace{-20pt}
\section{Conclusions and Discussion}
\input{Texs/6_conclusion}



%
%
\bibliographystyle{splncs04}
\bibliography{main}

\clearpage
\setcounter{section}{0}
\renewcommand\thesection{\Alph{section}}

\section{Additional Information of Omni6DPose}
\label{sec:add_info_of_omni6dpose}
\input{SuppTexs/09_supp_Omni6DPose}

\section{GenPose++ Implementation Details}
\label{sec:Genpose++_details}
\input{SuppTexs/11_supp_Genpose++_details}

\section{Visualization of {\SimData}}
\label{sec:SOPE_visualization}
\input{SuppTexs/10_supp_SOPE}

\section{Additional Experiments and Results}
\label{sec:add_experiments}
\input{SuppTexs/12_supp_add_experiments}

\nocite{ci2023gfpose}
\nocite{wu2024unidexfpm}
\nocite{wu2024unidexfpm}
\end{document}

%% file: Texs/notations.tex
\newcommand{\Yes}{\textcolor{green}{\ding{51}}}
\newcommand{\No}{\textcolor{red}{\ding{55}}}

\newcommand{\ModelName}{GenPose++}
\newcommand{\Dataset}{Omni6DPose}
\newcommand{\RealData}{ROPE}
\newcommand{\SimData}{SOPE}

\newcommand{\ClsNum}{149}
\newcommand{\SInsNum}{4162}
\newcommand{\SImgNum}{475K}
\newcommand{\SAnnoNum}{5M}
\newcommand{\RInsNum}{581}
\newcommand{\RImgNum}{332K}
\newcommand{\RAnnoNum}{1.5M}
\newcommand{\RVidNum}{363}

\newcommand{\obj}{O}
\newcommand{\img}{I}
\newcommand{\trans}{T}
\newcommand{\rot}{R}
\newcommand{\pose}{\vb*{p}}
\newcommand{\cand}{\hat{\pose}}

\newcommand{\dist}{p_{\text{data}}}
\newcommand{\energy}{\vb*{\Psi}_{\phi}}
\newcommand{\score}{\vb*{\Phi}_{\theta}}

\newcommand{\percent}{\delta}

\newcommand{\posespace}{\mathcal{P}}

\newcommand{\eps}{\epsilon}

\newcommand{\E}{\mathbb{E}}
\newcommand{\R}{\mathbb{R}}

\newcommand{\s}{\vb*{s}}

\newcommand{\Tref}[1]{Tabel~\ref{#1}}
\newcommand{\Fref}[1]{Figure~\ref{#1}}

\newcommand{\Sref}[1]{Section~\ref{#1}}

%% file: Texs/0_abstract.tex
6D Object Pose Estimation is a crucial yet challenging task in computer vision, suffering from a significant lack of large-scale datasets. This scarcity impedes comprehensive evaluation of model performance, limiting research advancements. Furthermore, the restricted number of available instances or categories curtails its applications.
To address these issues, this paper introduces \textbf{{\Dataset}}, a substantial dataset characterized by its diversity in object categories, large scale, and variety in object materials. {\Dataset} is divided into three main components: \textbf{{\RealData}} (\textbf{Real 6D Object Pose Estimation Dataset}), which includes {\RImgNum} images annotated with over {\RAnnoNum} annotations across {\RInsNum} instances in {\ClsNum} categories; \textbf{{\SimData}}(\textbf{Simulated 6D Object Pose Estimation Dataset}), consisting of {\SImgNum} images created in a mixed reality setting with depth simulation, annotated with over {\SAnnoNum} annotations across {\SInsNum} instances in the same {\ClsNum} categories; and the manually aligned real scanned objects used in both {\RealData} and {\SimData}. 
{\Dataset} is inherently challenging due to the substantial variations and ambiguities. 
To address this challenge, we introduce \textbf{{\ModelName}}, an enhanced version of the SOTA category-level pose estimation framework, incorporating two pivotal improvements: Semantic-aware feature extraction and Clustering-based aggregation.
Moreover, we provide a comprehensive benchmarking analysis to evaluate the performance of previous methods on this large-scale dataset in the realms of 6D object pose estimation and pose tracking. Additional demonstrations are available at \url{https://omni6dpose-pending.vercel.app}.

%% file: Texs/1_intro.tex
\begin{figure*}[tbp]
\centerline{\includegraphics[width=\textwidth]{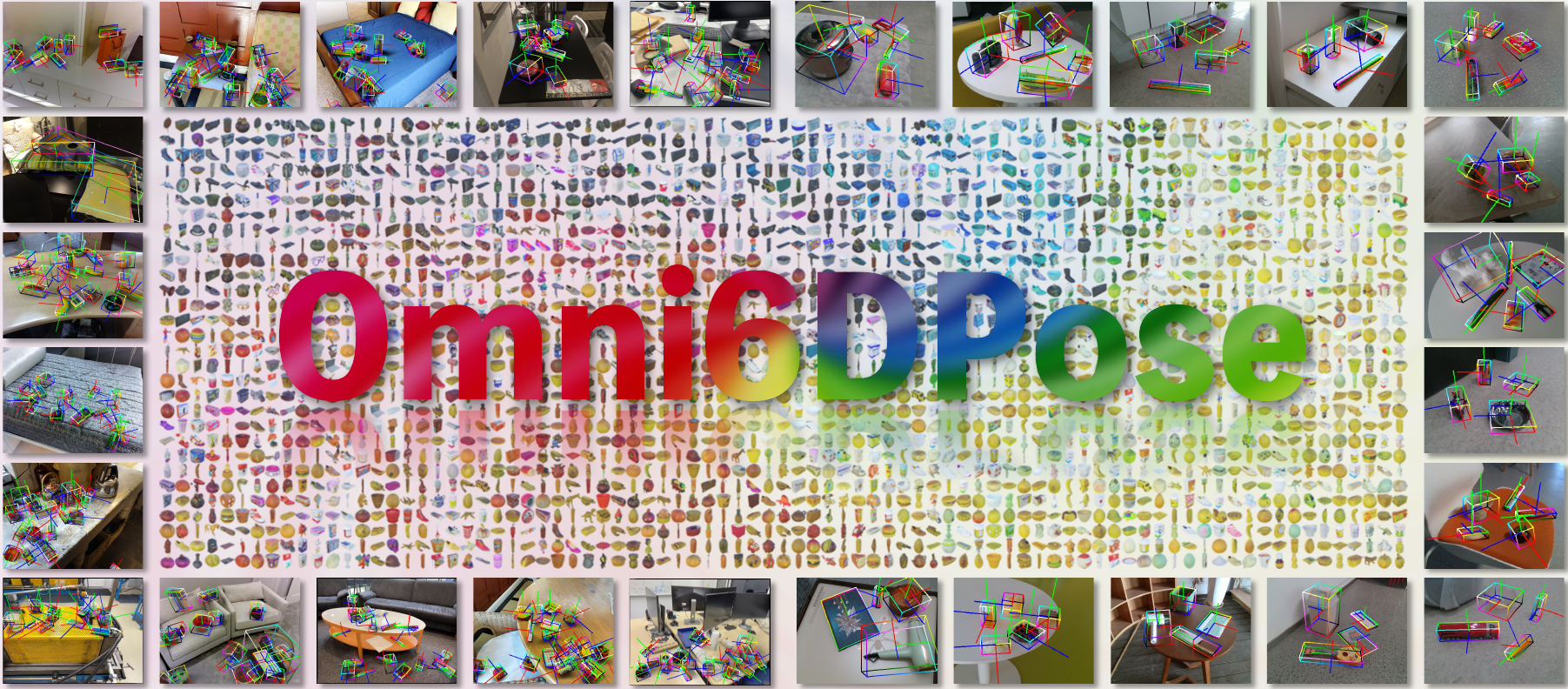}}
\vspace{-5pt}
\caption{
We introduce a universal 6D object pose estimation dataset, \textbf{\Dataset}. The middle section showcases some examples of canonically aligned objects from our dataset, with samples of {\SimData} depicted on the left and samples of {\RealData} on the right.
}
\vspace{-10pt}
\label{fig:teaser}
\end{figure*}

6D object pose estimation \cite{Wang_2019_CVPR, zhou2023drpose, liu2023prior} and pose tracking \cite{hff6d, 9568706} from single images is an essential task in computer vision, holding immense potential for applications in robotics \cite{an2023rgbmanip} and augmented reality/virtual reality (AR/VR)\cite{7368948}. Over recent decades, the domain has experienced significant advancements, primarily dominated by data-driven learning approaches. Analogous to the pivotal role of data in learning-based 2D foundation tasks, high-quality, comprehensive datasets are paramount in the context of 6D object pose estimation and tracking.

Today 6D object pose estimation is studied under two different lenses: instance-level and category-level. In instance-level settings, datasets such as Linemod~\cite{linemod}, YCB-Video~\cite{xiang2018posecnn}, and T-LESS~\cite{hodan2017tless} have gained widespread acceptance as benchmarks. These datasets are distinguished by their focus on detailed, individual object instances, thereby enabling algorithms to precisely learn and predict the poses of specific items. On the other hand, category-level pose estimation emphasizes generalization across different items within a particular object category. The NOCS~\cite{Wang_2019_CVPR} dataset stands out as the most widely used in the category-level object pose estimation field, providing a simulated dataset for training and a small-scale real-world dataset for evaluation. Despite their contributions to advancing the field, these datasets present limitations due to their small scale in terms of instances or categories. This results in two significant challenges:

\begin{enumerate}
\item It hampers comprehensive evaluation of different models' performance, limiting the development of research in this field.
\item It restricts the applicability of research findings across diverse domains, due to the limited variety of object instances or categories represented.
\end{enumerate}

To address the aforementioned challenges and drive advancements in this field, this paper introduces \textbf{\Dataset}, a universal 6D object pose estimation dataset characterized by its diversity in object categories, expansive scale, and variety in materials. {\Dataset} is segmented into three principal components: 1) \textbf{\RealData} (\textbf{Real 6D Object Pose Estimation Dataset}), which encompasses {\RImgNum} images annotated with over {\RAnnoNum} annotations across {\RInsNum} instances in {\ClsNum} categories; 2) \textbf{\SimData} (\textbf{Simulated 6D Object Pose Estimation Dataset}), comprising {\SImgNum} images generated in a mixed reality setting with depth simulation, furnished with over {\SAnnoNum} annotations spanning {\SInsNum} instances in the same {\ClsNum} categories. The mixed reality bridges the semantic sim2real gap, while the depth sensor simulation close the geometric sim2real gap; 3) the manually aligned, real scanned objects utilized in both {\RealData} and {\SimData}, enabling the generation of diverse downstream task data.

\begin{figure*}[tbp]
\begin{center}
\centerline{\includegraphics[width=\textwidth]{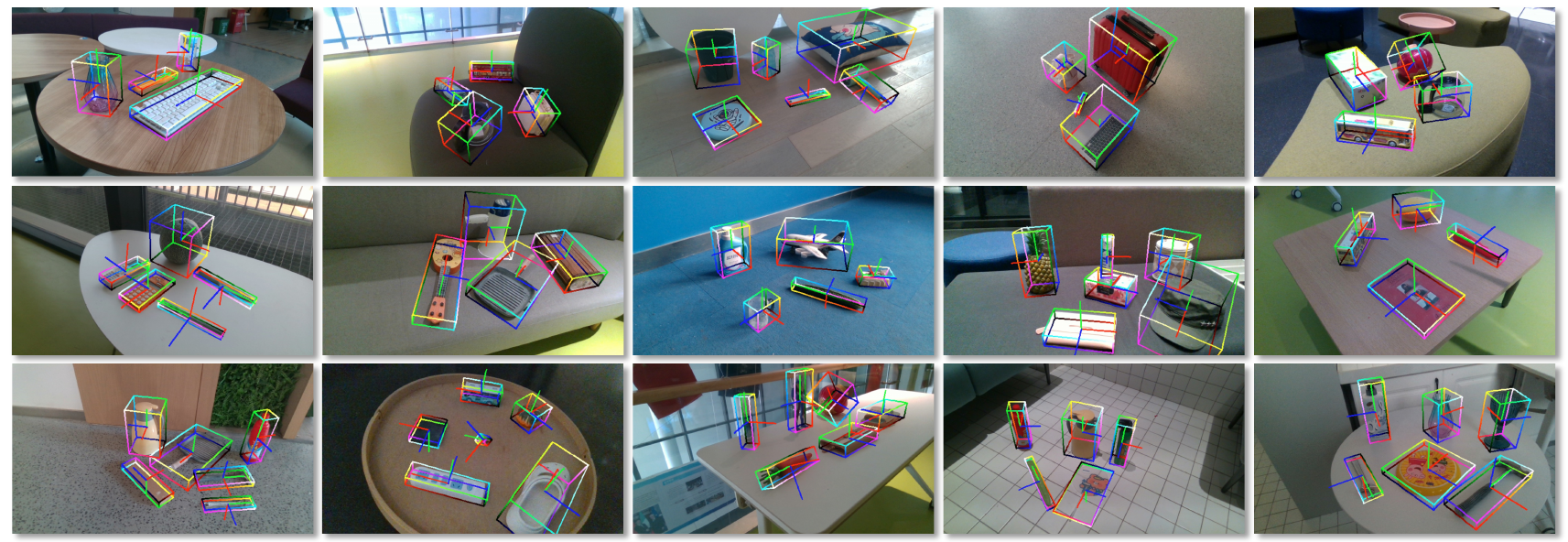}}
\vspace{-5pt}
\caption{
\textbf{{\RealData} dataset visualization.} In the figure, bounding boxes are colored according to the coordinates in the object's coordinate system.
}
\vspace{-30pt}
\label{fig:ROPE_visualization}
\end{center}
\end{figure*}

\textbf{\Dataset} poses inherent challenges due to its considerable variations, diverse materials, and inherent ambiguities, which reflect the complexities encountered in real-world applications. \Fref{fig:ROPE_visualization} illustrates some examples from {\RealData}. 
To tackle these issues, we introduce \textbf{{\ModelName}}, which incorporates GenPose~\cite{zhang2023genpose} with two pivotal improvements: Semantic-aware feature extraction and Clustering-based aggregation, tailored specifically to the nuances of the {\Dataset} in question.
Furthermore, as a Universal 6D object pose estimation dataset, this paper also offers a comprehensive benchmarking analysis to assess the performance of existing methods on category-level 6D object pose estimation and pose tracking. 
We summarize our contributions as follows:
\begin{enumerate}
\item We present \textbf{\Dataset}, a comprehensive 6D object pose estimation dataset with extensive categories, instance diversity, and material variety.
\item We propose a real data collection pipeline and a simulation framework for generating synthetic data with low semantic and geometry sim2real gaps.
\item We introduce \textbf{\ModelName} for category-level 6D object pose estimation and tracking, demonstrating SOTA performance on {\Dataset}.
\end{enumerate}

%% file: Texs/2_related_work.tex
For 6D object pose estimation, there are two main branches: instance-level and category-level. Instance-level estimation is tested on seen objects or test on unseen objects with known CAD model, while category-level estimation is tested on unseen instances of known categories without CAD model.
In this section, we review and compare existing datasets to our large-scale category-level dataset and review relevant algorithms for category-level pose estimation and tracking.

\begin{table}[t]
\caption{This table compares datasets for 6D object pose estimation, focusing on object category count, reality of the data, data modalities (RGB, Depth, IR), and object attributes such as quantity, CAD model availability, and inclusion of transparent and specular objects. It also details video characteristics by number and marker presence, along with image and annotation counts. `Wild6D$^*$' refers specifically to the test split of the Wild6D dataset, as the training data does not provide annotations. The symbol `-' indicates the absence of a particular feature within the dataset.}
\vspace{-10pt}
\resizebox{\textwidth}{!}{
\input{Tables/data_statistics}

}
\vspace{-15pt}
\label{table: dataset_comparision}
\end{table}

\vspace{-10pt}
\subsection{6D Object Pose Estimation Datasets}
Following the outlined branches of 6D object pose estimation, we have reviewed datasets corresponding to both instance-level and category-level 6D object pose estimation. A comparative analysis of these datasets is provided in \Tref{table: dataset_comparision}. 
\subsubsection{Instance-level 6D object pose estimation dataset.}
LineMod~\cite{linemod} is one of the most used datasets, providing non-temporal RGB-D images and ground truth pose annotations. YCB-Video~\cite{xiang2018posecnn} provides RGB-D videos and annotations, enabling pose-tracking approaches. T-LESS~\cite{hodan2017tless} features texture-less objects with symmetries and mutual similarities. StereoOBJ-1M~\cite{liu2021stereobj1m} achieves a leap in dataset scale and features transparent and reflective objects. While these datasets are extensive in terms of image and annotation count, they are limited in the diversity of instances they cover. For example, StereoOBJ-1M comprises 339K frames and 1.5M annotations, yet it includes only 18 unique object instances.

\vspace{-10pt}
\subsubsection{Category-level 6D object pose estimation dataset.}
NOCS~\cite{Wang_2019_CVPR} provides the first benchmark in category-level pose estimation. Wild6D~\cite{fucategory} addresses the scalability issue of datasets by leveraging unlabeled and synthetic data. PhoCal~\cite{wang2022phocal} focuses on photometrically challenging objects and proposes a high-accuracy annotating pipeline. HouseCat6D~\cite{jung2023housecat6d} offers diverse scenes, viewpoints, and grasping annotations. 
However, these datasets cover only a limited number of categories. Even the most extensive dataset, HouseCat6D, includes merely ten categories.
Our datasets, {\SimData} and {\RealData}, set new benchmarks by offering the widest range of categories and featuring objects with diverse materials, thereby enhancing dataset diversity and realism for pose estimation research.

\subsection{6D Object Pose Estimation and Tracking Algorithm}

\subsubsection{Category-leval 6D object pose estimation.}
Category-level 6D object pose estimation aims to estimate unseen instance poses within the same object category. 
NOCS introduces a normalized object coordinate space for pose prediction without CAD models, while SPD~\cite{wang2021category} and SGPA~\cite{chen2021sgpa}
utilize category-level priors for enhanced estimation. 
HS-Pose~\cite{Zheng_2023_CVPR} extends 3D-GC~\cite{Lin_2020_CVPR} for improved feature extraction from point clouds. 
IST-Net~\cite{liu2023prior} transforms features between camera and world spaces implicitly, without relying on 3D priors, surpassing previous methods.
However, these techniques, mainly regression-based, require ad-hoc designs for symmetric objects. GenPose~\cite{zhang2023genpose} addresses it by generatively modeling the pose distribution, eliminating the need for symmetry considerations. Yet, GenPose neglects RGB semantic information, which is increasingly vital as object category scales grow. Additionally, its energy-based aggregation algorithm fails with discontinuous multimodal distributions.
We introduce GenPose++, which incorporates a 2D foundation model to leverage RGB semantics, improving generalization, and introduces an aggregation module to handle discrete multimodal distributions, addressing the limitations of GenPose.

\vspace{-10pt}
\subsubsection{6D object pose tracking.}
This paper is situated within the domain of category-level 6D object pose tracking and model-free object tracking. BundleTrack~\cite{wen2021bundletrack} pioneers model-free tracking by leveraging multi-view feature detection for tracking unseen objects without 3D models. CAPTRA~\cite{weng2021captra} enhances articulated pose tracking through recursive updates for better temporal consistency. CATRE~\cite{liu2022catre} aligns partially observed point clouds to abstract shape
priors for relative transformations and pose estimation. GenPose, adopting a generative approach, effectively addresses the challenge of pose ambiguities in symmetric objects. Together, these approaches underscore the evolving landscape of object pose tracking, highlighting both the progress and the diversity of strategies being explored.

%% file: Tables/data_statistics.tex
{
\newcommand{\A}{\multirow{2}{*}{Dataset}}
\newcommand{\B}{\multirow{2}{*}{Cat.}}
\newcommand{\C}{\multirow{2}{*}{Real}}
\newcommand{\D}{\multicolumn{3}{c}{Modality}}
\newcommand{\EE}{\multicolumn{4}{c}{Object}}
\newcommand{\F}{\multirow{2}{*}{Marker-free}}
\newcommand{\J}{\multirow{2}{*}{Vid.}}
\newcommand{\G}{\multirow{2}{*}{Img.}}
\newcommand{\I}{\multirow{2}{*}{Anno.}}
\newcommand{\colour}{\rowcolor{gray!15}}

\hskip-1.0cm\begin{tabular}{ccc|ccc|cccc|cc|cc}
\toprule
\A                                      & \B   & \C    & \D                           & \EE                              & \F       & \J         & \G         & \I         \\ \cmidrule{4-6} \cmidrule{7-10}
                                        &      &       & RGB     & Depth     & IR     & Num.     & CAD  & Trans. & Spec. &          &            &            &            \\ \midrule \midrule
CAMERA~\cite{Wang_2019_CVPR}            & 6    & \No   & \Yes    & \Yes      & \No    & 1085     & \Yes & \No    & \No   & \Yes     & -          &300K        &4M          \\
\colour SOPE(Ours)                      & 149  & \No   & \Yes    & \Yes      & \Yes   & \SInsNum & \Yes & \Yes   & \Yes  & \Yes     & -          &\SImgNum    &\SAnnoNum   \\ \midrule \midrule
YCB-Video~\cite{xiang2018posecnn}       & -    & \Yes  & \Yes    & \Yes      & \No    & 21       & \Yes & \No    & \No   & \Yes     & 92         &133K        &    598K    \\
T-LESS~\cite{hodan2017tless}            & -    & \Yes  & \Yes    & \Yes      & \No    & 30       & \Yes & \No    & \No   & \No      & 20         &48K         &48K         \\
Linemod~\cite{linemod}                  & -    & \Yes  & \Yes    & \Yes      & \No    & 15       & \Yes & \No    & \No   & \No      & -          &18K         & 15K        \\
StereoOBJ-1M~\cite{liu2021stereobj1m}   & -    & \Yes  & \Yes    & \No       & \No    & 18       & \Yes & \Yes   & \Yes  & \No      & 182        &393K        &1.5M        \\ \midrule
REAL275~\cite{Wang_2019_CVPR}           & 6    & \Yes  & \Yes    & \Yes      & \No    & 42       & \Yes & \No    & \No   & \No      & 18         &8K          &35K      \\
PhoCaL~\cite{wang2022phocal}            & 8    & \Yes  & \Yes    & \Yes      & \No    & 60       & \Yes & \Yes   & \Yes  & \Yes     & 24         &3.9K        &91K         \\
HouseCat6D~\cite{jung2023housecat6d}    & 10   & \Yes  & \Yes    & \Yes      & \No    & 194      & \Yes & \Yes   & \Yes  & \Yes     & 41         &23.5K       &160K        \\
Wild6D$^*$~\cite{fucategory}            & 5    & \Yes  & \Yes    & \Yes      & \No    & 162      & \No  & (\Yes) & \No   & \Yes     & 486        &10K         &10K         \\
\colour ROPE(Ours)                      & 149  & \Yes  & \Yes    & \Yes      & \Yes   & \RInsNum & \Yes & \Yes   & \Yes  & \Yes     & \RVidNum   &\RImgNum    &\RAnnoNum   \\ \bottomrule
\end{tabular}
}

%% file: Texs/3_data_creation.tex
This paper introduces a rich variety of object categories, a large-scale, and materially diverse dataset for real 6D object pose estimation, named {\RealData}. And, a simulated dataset, SOPE, synthesizing with mixed reality and featuring depth simulation, is provided for training. \Sref{sec: 3D Scanniing} will discuss the collection and alignment of 3D objects. \Sref{sec: ROPE creation} will cover the acquisition and labeling of the {\RealData} dataset. \Sref{sec: SOPE Synthesis} will explain the generation of the {\SimData} dataset.

\begin{figure*}[tbp]
\begin{center}
\centerline{\includegraphics[width=\textwidth]{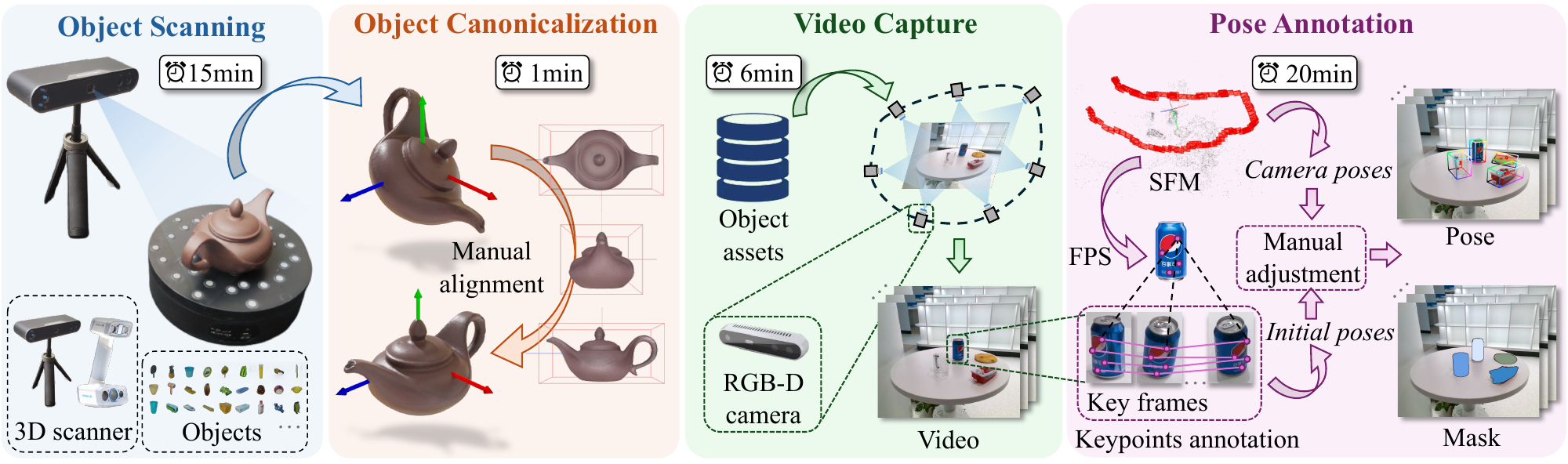}}
\vspace{-5pt}
\caption{
\textbf{\RealData} dataset collection and annotation. (1) Object scanning, where high-precision industrial scanners are used to acquire the CAD models of objects; (2) Object canonicalization, involving the alignment of each object category to the canonical space; (3) Video capture, capturing video sequences in varied scenarios with a depth camera; and (4) Pose annotation, calculating camera poses through Structure from Motion (\textbf{SFM}), further utilizing Farthest Point Sampling (\textbf{FPS}) to select keyframes for keypoint annotation, and performing bundle adjustment to derive initial object pose values, which are then manually refined to obtain more precise annotations.
}
\label{fig: ROPE}
\vspace{-30pt}
\end{center}
\end{figure*}

\subsection{3D Object Collection and Alignment}
\label{sec: 3D Scanniing}

Universal 6D object pose estimation relies on a comprehensive set of objects. We selected 149 categories of everyday objects, all reconstructed with high-precision scanners, and categorized them into two sets: {\SimData} for simulated data and {\RealData} for real-world scenes. {\SimData} primarily includes objects from sources like OmniObject3D~\cite{wu2023omniobject3d}, PhoCal~\cite{wang2022phocal}, and GoogleScan~\cite{downs2022google}, alongside a subset from our scans, totaling 5,000 instances. {\RealData} consists of 580 instances we reconstructed using industrial scanners. Importantly, while most {\SimData} objects are from public datasets, manual category-level pose alignment is necessary. 
For object reconstruction, as shown in \Fref{fig: ROPE}, we use two professional scanners, EinScan H2\footnote{\url{https://www.einscan.com/}} and Revopoint POP 3 \footnote{\url{https://www.revopoint3d.com/}}  for objects in different scales. The scanning time depends on object features: it took about 15 minutes for small, simple, Lambertian items like a mouse, and up to an hour for complex, large, or non-Lambertian items like a transparent mug.
Finally, we constructed a specialized annotation tool for manually aligning objects in the same category to the category-level canonical space, with each alignment taking roughly one minute.

\vspace{-10pt}
\subsection{{\RealData} Acquisition and Annotation}
\label{sec: ROPE creation}

The ROPE dataset was systematically acquired utilizing the RealSense D415 imaging device, encompassing scenarios with 2 to 6 distinct objects and video sequences extending from 762 to 1,349 frames. The integrity and utility of the dataset are underpinned by the precision of object pose annotations, which present notable challenges, chiefly:

\begin{enumerate}
    \item Derivation of relative camera poses, denoted as $T_c = \{(R_i, t_i)\}_{i=1}^n$, where each pair $(R_i, t_i) \in \text{SE}(3)$, signifying the transformation from the camera space to the world space for the \(i^{th}\) frame, with \(n\) symbolizing the aggregate frame count within the video sequence.
    \item Procurement of high-accuracy object poses, represented as $T_o = \{(R, t)\}$, where the pair $(R, t) \in \text{SE}(3)$, delineating the transformation from the object space the camera space for any selected frame.
\end{enumerate}

With \(T_c\) and \(T_o\) known, it is possible to automate the generation of all annotations within the dataset. 
Addressing these challenges, we propose a marker-free annotation system. Previous datasets for calculating relative camera pose rely on markers, like NOCS~\cite{Wang_2019_CVPR}, or external robot arms for indirect calculation, such as PhoCal~\cite{wang2022phocal}, limiting scene diversity. As shown in \Fref{fig: ROPE}, to enable marker-free annotation in open scenes, we consider it a structure-from-motion (SfM) problem, utilizing intrinsic scene features to optimize camera poses $T_c$ using bundle adjustment techniques. This approach aims to solve the optimization problem:

\begin{equation}
    \min_{T_c} \sum_{i=1}^{n} \sum_{j \in \mathcal{P}_i} \| \pi(R_i X_j + t_i) - x_{ij} \|^2
\end{equation}

where $\pi$ denotes the camera projection function, $X_j$ represents the 3D points in the world space, and $x_{ij}$ corresponds to the 2D projection of $X_j$ in the $i^{th}$ camera frame, with $\mathcal{P}_i$ being the set of point correspondences in frame $i$.

For object pose $T_o$ annotations, previous methods consider only a single frame, leading to inaccuracies due to the lack of multi-viewpoint constraints. To overcome this, we introduce a two-stage object pose annotation process: in the first stage, keyframes are sampled from SfM results using Farthest Point Sampling (FPS), and 2D-3D keypoint pairs on these keyframes are annotated, providing initial object pose by minimizing the reprojection error:

\begin{equation}
    \min_{R, t} \sum_{k \in \mathcal{K}} \| \pi(R X_k + t) - x_k \|^2
\end{equation}

where $X_k$ are the 3D keypoints of the object, $x_k$ are their corresponding 2D annotations in the image, and $\mathcal{K}$ is the set of all keypoint correspondences. In the second stage, these initial poses are manually fine-tuned based on the object's projection across all keyframes.

\vspace{-10pt}
\subsection{{\SimData} Synthesis}
\label{sec: SOPE Synthesis}

\begin{figure*}[tbp]
\begin{center}
\centerline{\includegraphics[width=\textwidth]{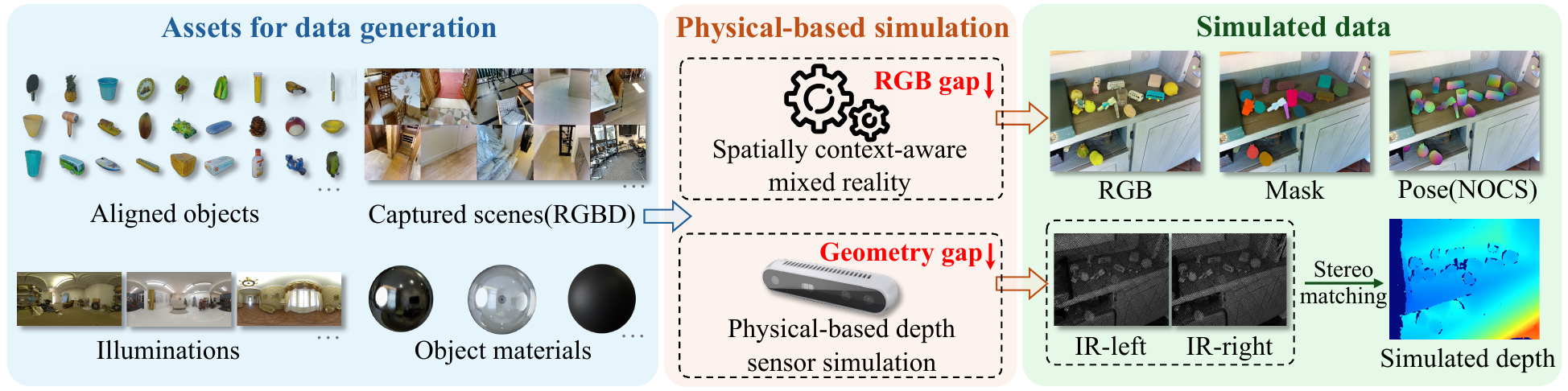}}

\vspace{-5pt}
\caption{
\textbf{\SimData} synthesis, utilizing mixed reality to bridge the RGB sim2real gap and physical-based depth sensor simulation to minimize the geometric sim2real gap.
}
\label{fig:SOPE}
\vspace{-40pt}
\end{center}
\end{figure*}

ROPE represents a comprehensive benchmark in category-level object pose estimation by scaling up the diversity and number of object categories to unprecedented levels, encompassing a wide range of materials. This diversity presents new challenges for network training data due to the higher demands on the dataset's scale and diversity. Collecting a larger real-world dataset would be prohibitively expensive and unlikely to ensure sufficient diversity.

To bridge the sim2real gap, which is pronounced when using synthetic data, either in RGB or geometry, this paper proposes a novel method based on mixed reality with depth simulation for synthetic data generation. Specifically, as demonstrated in \Fref{fig:SOPE}, we employ mixed reality~\cite{Wang_2019_CVPR} techniques to generate RGB data, thereby reducing the RGB sim2real gap. In parallel, we simulate the mechanism of structured light depth sensors within blender~\cite{dai2022domain}. This involves rendering infrared (IR) images and applying stereo matching to produce synthetic depth maps, effectively narrowing the geometry's sim2real gap.

During data generation, we implement domain randomization for illumination and object materials to further enhance the dataset's diversity. All the background images are sourced from public datasets, including 19,658 images from MatterPort3D~\cite{chang2017matterport3d}, 2,572 from ScanNet++~\cite{yeshwanth2023scannet++}, and 540 from IKEA~\cite{Wang_2019_CVPR}. To the best of our knowledge, this is the first simulated dataset that uses a context-aware mixed reality approach combined with physical-based depth sensor simulation for object pose estimation tasks.

\begin{figure*}[tbp]
\begin{center}
\centerline{\includegraphics[width=\textwidth]{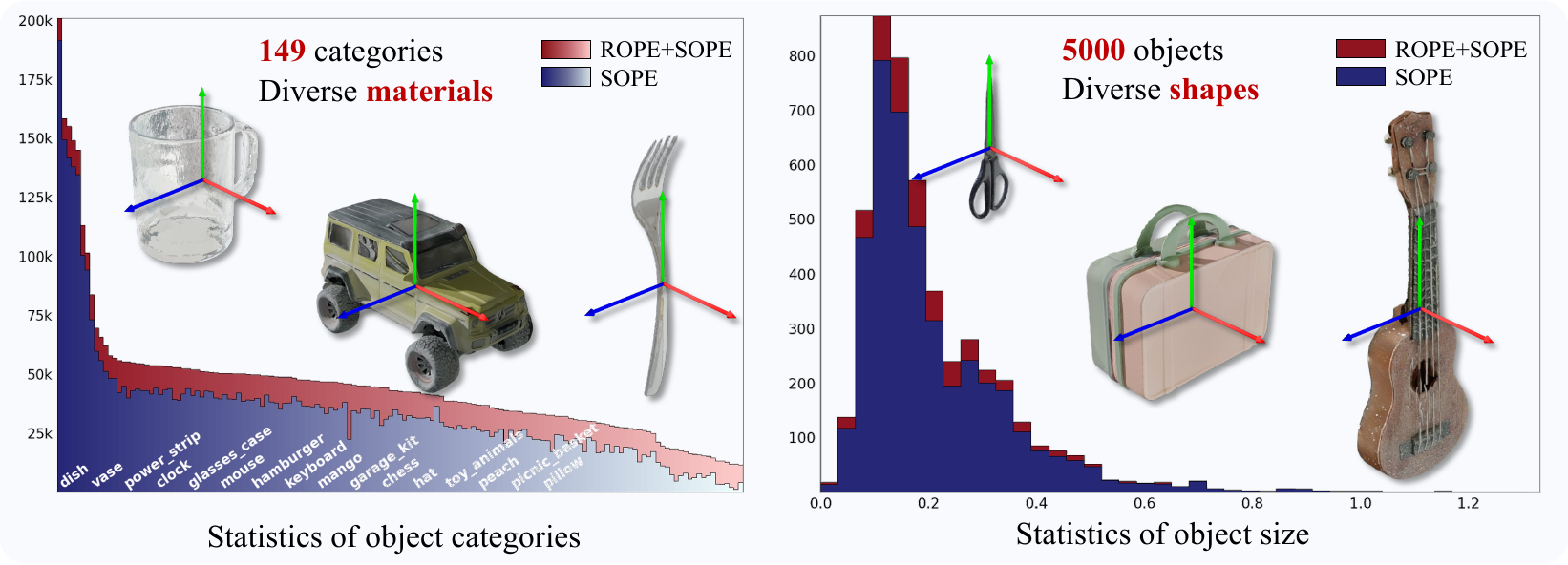}}

\vspace{-5pt}
\caption{
\textbf{\Dataset} statistics, showcasing the dataset distribution. Left: Category distribution, highlighting 149 categories and diverse materials. Right: Object size distribution across 5000 objects, illustrating diversity in shapes.
}
\label{fig:data_stats}
\vspace{-30pt}
\end{center}
\end{figure*}

\vspace{-10pt}
\subsection{Dataset Statistics}

\vspace{-5pt}
\subsubsection{Object Category Statistics}
The comprehensive distribution of object category and size are both demonstrated
in Figure \ref{fig:data_stats}.
Most of the categories possess $\ge 25$K pose annotations in the
{\SimData} dataset, providing sufficient training opportunities. Categories
containing objects with diverse and challenging material options 
(e.g., transparent or specular materials) are equipped with apparently more
data generation, such as dishes, cups, bottles, bowls, mugs, etc.

\vspace{-10pt}
\subsubsection{Object Size Statistics}
As shown in Figure \ref{fig:data_stats}, the objects in our dataset span a wide range of sizes. The majority of the objects are approximately 0.1 meters in length along the diagonal of their bounding boxes, with the largest objects exceeding 1 meter.

%% file: Texs/4_method.tex
\begin{figure*}[tbp]
\begin{center}
\centerline{\includegraphics[width=\textwidth]{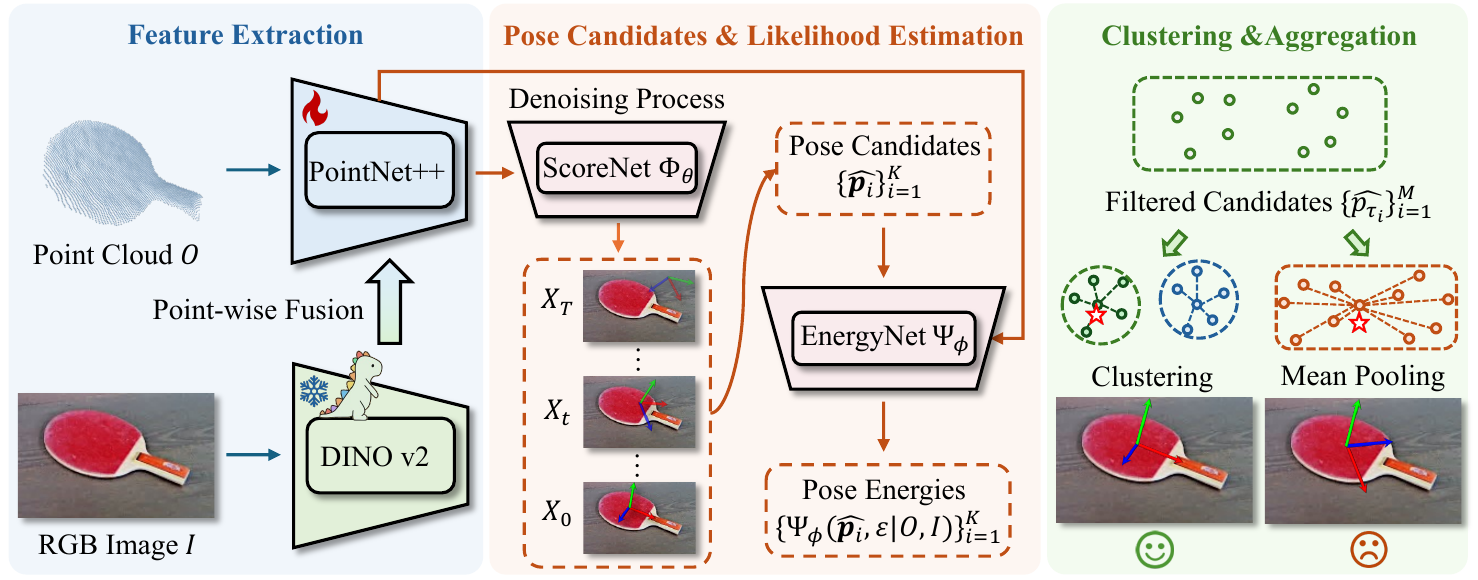}}
\caption{
\textbf{Overview of {\ModelName}}. GenPose++ employs segmented point clouds and cropped RGB images as inputs, utilizing PointNet++ for extracting object geometric features. Concurrently, it employs a pre-trained 2D foundation backbone, DINO v2, to extract general semantic features. These features are then fused as the condition of a diffusion model to generate object pose candidates and their corresponding energy. Finally, clustering is applied to address the aggregation issues associated with the multimodal distribution of poses for objects exhibiting non-continuous symmetry, such as boxes, effectively resolving the pose estimation challenge.
}
\label{fig:method}

\vspace{-30pt}
\end{center}
\end{figure*}




Given the {\Dataset}, one naturally ponders the optimal technical approach for large-scale category-level pose estimation.
The recently introduced state-of-the-art category-level 6D pose estimation technique, GenPose~\cite{zhang2023genpose}, offers a promising avenue by employing a diffusion-based probabilistic method~\cite{song2020score, song2019generative}. In contrast, the diffusion model has demonstrated remarkable efficacy across various high-dimensional domains with extensive training data~\cite{ci2023gfpose, tian2024robokeygen, zeng2023distilling, wu2024learning}.


Expanding on this groundwork, our study delves further into the probabilistic approach, presenting an enhanced iteration of GenPose, named {\ModelName}. {\ModelName} integrates two crucial enhancements: Semantic-aware feature extraction (see Fig.\ref{fig:method} (a)) and Clustering-based aggregation (as shown in Fig.\ref{fig:method} (c)). The subsequent sections will detail the three primary stages of the {\ModelName} pipeline. Moreover, given the estimated 6D pose, GenPose++ provide a additional regression network to predict the 3D scale of the object.

\vspace{-10pt}
\subsection{Training Semantic-aware Score and Energy Networks}
The learning agent is trained on our paired dataset $\mathcal{D} = \{ (\pose_k, \obj_k, \img_k)\}_{k=1}^n$, where $\pose_k \in \text{SE}(3)$, $\obj_k \in \mathbb{R}^{3\times N}$, and $\img_k \in \mathbb{R}^{3\times H \times W}$ denote a 6D pose, a partially observed 3D point cloud with $N$ points, and a cropped RGB image with $H\times W$ resolution, respectively.
Given an unseen object with point cloud $\obj^*$ and RGB image $\img^*$, the goal is to recover the corresponding ground-truth pose $\pose^*$.

Following GenPose, we initially train a score network $\score: \R^{|\posespace|} \times \R^1 \times \R^{3\times N} \times \R^{3\times H \times W}  \rightarrow \R^{|\posespace|}$ and an energy network $\energy: \R^{|\posespace|} \times \R^1 \times \R^{3\times N} \times \R^{3\times H \times W} \rightarrow \R^{1}$ from the dataset $\mathcal{D}$ using the denoising score-matching objective~\cite{denosingScoreMatching}:
\begin{equation}
\begin{aligned}
   \E_{
   t\sim \mathcal{U}(\eps, 1)}
   \left\{
   \lambda(t)\E_{
   \pose(0) \sim \dist(\pose(0)|\obj, \img), 
   \atop
   \pose(t) \sim \mathcal{N}(\pose(t);\pose(0), \sigma^2(t)\mathbf{I})
   }
   \left[ \left\Vert\s(\pose(t), t | \obj, \img)  - \frac{\pose(0) - \pose(t)}{\sigma(t)^2} \right\Vert_2^2 \right] \right\} 
\end{aligned}
\label{eq:score_matching_loss}
\end{equation}
The training loss of the score and energy network can be obtained by replacing $\s(\pose(t), t | \obj, \img)$ in Eq.~\ref{eq:score_matching_loss} with $\score(\pose(t), t | \obj, \img)$ and $\nabla_{\pose} \energy^*(\pose(t), t| \obj, \img)$, respectively.

Unlike GenPose, our score and energy network are semantic-aware, as both networks are conditioned on an RGB image to incorporate semantic cues for pose estimation. 
To fuse the features extracted from the image and point cloud, we encode the RGB image $\text{img}$ and point cloud $\text{obj}$ using the pre-trained feature extractors from DINOv2~\cite{oquab2023dinov2} and PointNet++\cite{qi2017pointnet++}, respectively. Then, we concatenate these features together in a pointwise manner similar to\cite{wang2019densefusion}.

\vspace{-10pt}
\subsection{Candidates Generation and Outlier Removal}
Following GenPose, we subsequently sample pose candidates $\{\cand_i \}_{i=1}^K$ by solving the following \textit{Probability Flow}~(PF) ODE~\cite{song2020score} constructed by the score network $\score$ from $t=1$ to $t=\eps$:
\begin{equation}
\begin{aligned}
    \frac{d\pose}{dt} = - \sigma(t)\dot{\sigma}(t)\score(\pose(t), t | \obj, \img)
\label{eq:reverse_sde}
\end{aligned}
\end{equation}
where $\pose(1) \sim \mathcal{N}(\mathbf{0}, \sigma_{\text{max}}^2\mathbf{I})$, $\sigma(t) = \sigma_{\text{min}}(\frac{\sigma_{\text{max}}}{\sigma_{\text{min}}})^t$, $\sigma_{\text{min}} = 0.01$ and $\sigma_{\text{max}} = 50$.

To remove the outliers in candidates, we sort the candidates into a sequence $ \cand_{\tau_1} \succ \cand_{\tau_2} ... \succ  \cand_{\tau_K}$ where:
\begin{equation}
\begin{aligned}
\cand_{\tau_{i}} \succ \cand_{\tau_{j}} \iff \energy(\cand_{\tau_{i}}, \eps | \obj) > \energy(\cand_{\tau_{j}}, \eps|\obj)
\end{aligned}
\label{eq:energy_distillation_loss}
\end{equation}
Then, we filter out the last $1 - \percent$\% candidates and obtain  $\cand_{\tau_1} \succ \cand_{\tau_2} ... \succ  \cand_{\tau_{M}}$  where $\delta = 40\%$ is a hyper parameter and $M = \lfloor \percent\cdot K \rfloor$.

\vspace{-5pt}
\subsection{Clustering-based Aggregation}
\vspace{-5pt}
In this section, we aggregate the remaining candidates $\{\cand_{\tau_i} = (\hat{\trans}_{\tau_i}, \hat{\rot}_{\tau_i})\}_{i=1}^{M}$ to obtain the final results. GenPose achieves this by simply mean-pooling the filtered candidates. However, this strategy will encounter a severe \textit{mean-mode issue} when the object possesses discrete symmetrical properties.
As illustrated in Fig.~\ref{fig:method}, a ping pong paddle has two symmetric ground truth poses~(modes). Since the score network has encountered both modes during training, an optimally trained score network will likely output candidates around both modes. Simply mean-pooling these candidates will yield the average of the two modes, known as the `mean mode', which will deviate from both modes.

To mitigate this issue, we introduce a clustering-based aggregation mechanism. We employ DBSCAN~\cite{ester1996density} to cluster the candidates. It identify dense regions in the data space, forming clusters based on the density of data points and effectively separating noise from meaningful patterns, without the need to specify the number of clusters. This is achieved through dynamic determination of cluster quantities based on distance threshold ($\varepsilon$) and density threshold ($\text{MinPts}$). For instance, in our empirical setting, we set $\varepsilon\approx 0.45\text{rad}$ and $\text{MinPts}=5$.
After clustering the candidates, we select the cluster with the largest number of objects and get the mean-pooling result as the final estimation following GenPose. 




%% file: Texs/5_experiments.tex
\vspace{-5pt}
\subsection{Category-Level 6D Object Pose Estimation}

\subsubsection{Metric}

In prior studies, metrics such as the mean average precision (mAP) for 3D bounding box IoU, and the mean average precision (mAP) for objects with translation errors less than $m$ cm and rotation errors less than $n^\circ$ have been commonly used~\cite{Wang_2019_CVPR}. These metrics assess the performance of pose estimation models which typically involve two steps: 1) instance-level object segmentation, and 2) estimating object poses from these detections. However, these pose estimation metrics are influenced by the detection model's performance.
To concentrate on evaluating the precision of model pose estimation, we assume ground truth instance segmentation is known and propose the following two metrics:

\begin{itemize}
    \item \textbf{AUC@IoU$_n$}: This metric assesses the accuracy of predicted 3D bounding boxes, calculated via the Area Under the Curve (AUC) from various Intersection over Union (IoU) thresholds starting at $n$. In our study, we utilize \text{AUC@IoU$_{25}$}, \text{AUC@IoU$_{50}$}, and \text{AUC@IoU$_{75}$} as the benchmarks.
    
    \item \textbf{VUS@$n^{\circ}m$cm}: This metric offers a detailed analysis of 6D pose estimation accuracy, derived from the Volume Under Surface (VUS) across ranges of rotational (up to $n^{\circ}$) and translational (up to $m$cm) error thresholds. VUS aggregates the accuracy of pose predictions within set boundaries. In this paper, we apply \text{VUS@$5^{\circ}2$cm}, \text{VUS@$5^{\circ}5$cm}, \text{VUS@$10^{\circ}2$cm}, and \text{VUS@$10^{\circ}5$cm} for comprehensive performance assessment.
\end{itemize}

\vspace{-10pt}
\subsubsection{Baselines}
We evaluate five category-level pose estimation methods: NOCS~\cite{Wang_2019_CVPR}, SGPA~\cite{chen2021sgpa}, HS-Pose~\cite{Zheng_2023_CVPR}, 
IST-Net~\cite{liu2023prior} and GenPose~\cite{zhang2023genpose}.
Except for NOCS, which conducts both object detection and pose estimation as a whole, all methods are equipped with ground truth detection results.
For SGPA, the prior point cloud of each category is constructed by randomly selecting an object from the training dataset.
Considering that previous methods' augmentations for symmetry properties are only applicable to specific object categories within the NOCS dataset and not suitable for all object categories in {\Dataset}, data augmentation for object symmetry is disabled during training for all baseline methods. All methods are trained on {\SimData} and directly test on {\RealData}.

\begin{table}[t]
    \centering
    \caption{
        \textbf{Quantitative comparison of category-level object pose estimation on {\RealData}.} $\uparrow$ represents a higher value indicating better performance, while $\downarrow$ represents a lower value indicating better performance. \textbf{Prior-free} indicates whether the method requires category prior information. The `-' indicates that GenPose does not predict the object scale.
    }
    
\vspace{-5pt}
    \resizebox{\textwidth}{!}{
        \input{Tables/cat_level_baselines}

    }
    \label{table:baselines}
    
\vspace{-10pt}
\end{table}

\begin{figure*}[t]
    \begin{center}
    \centerline{\includegraphics[width=\textwidth]{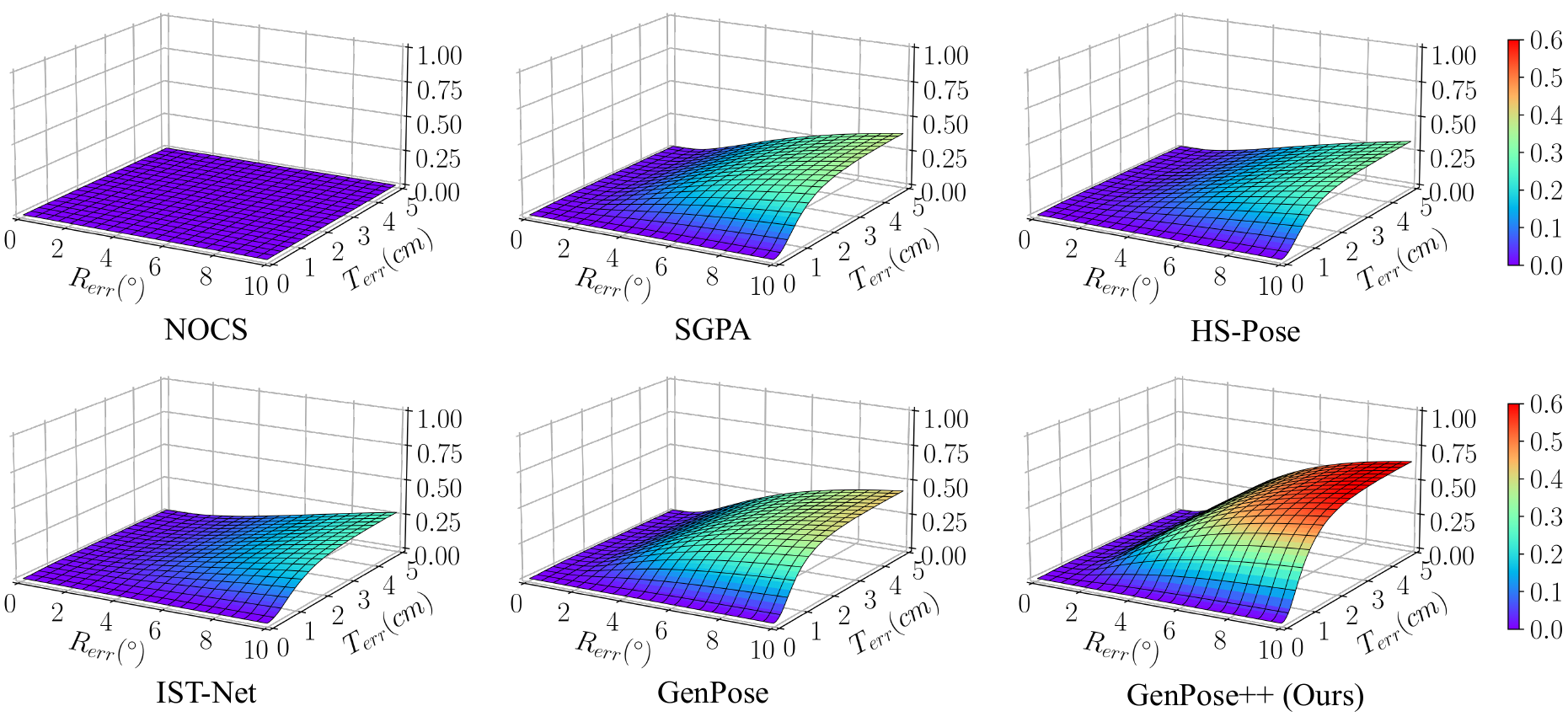}}
    
\vspace{-5pt}
    \caption{
        \textbf{Qualitative comparison with baselines on {\RealData} dataset.}
    }
    \label{fig:baseline_VUS}
    
\vspace{-35pt}
    \end{center}
\end{figure*}

\vspace{-10pt}
\subsubsection{Results and Analysis.}
\begin{figure*}[t]
    \begin{center}
    \centerline{\includegraphics[width=\textwidth]{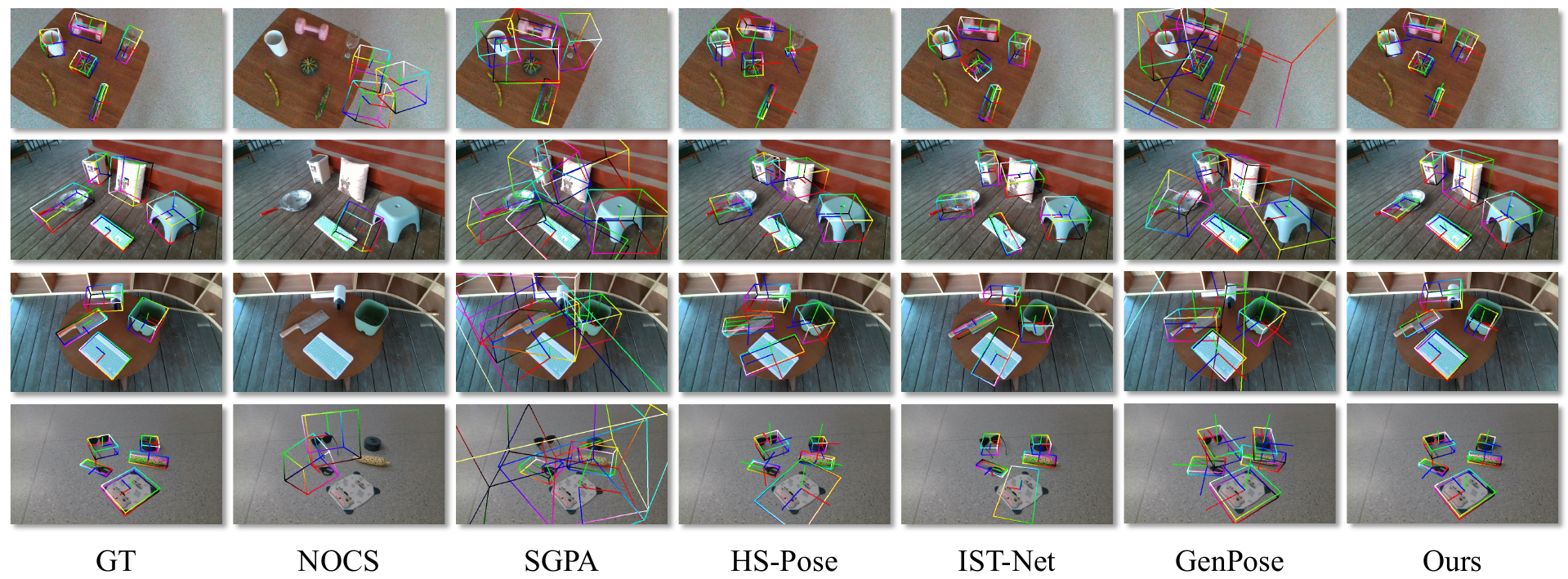}}
    
\vspace{-5pt}
    \caption{
        \textbf{Qualitative comparison with baselines on {\RealData} dataset.}
    }
    \label{fig:baseline_visualization}
    \end{center}
    
\vspace{-30pt}
\end{figure*}

In \Tref{table:baselines}, we present the quantitative evaluation results of previous methods compared to {\ModelName} on the {\RealData} dataset. Overall, generative methods continue to dominate in the performance evaluation on {\RealData}. The VUS surface depicted in \Fref{fig:baseline_VUS} provides a more detailed reflection of the performance of each model.
Unlike deterministic approaches, generative methods can handle ambiguity without any specialized design requirements. Moreover, these methods directly generate the distribution of object poses, eliminating the need for depth map-based pose fitting. This approach is particularly advantageous for challenging material types, such as transparent or reflective objects, where structured-light depth cameras tend to introduce significant noise, severely impacting pose fitting accuracy.
Furthermore, the NOCS method does not demonstrate effective performance on the {\RealData} dataset, leading to the supposition that methods relying solely on RGB information to predict the shape of an object in the canonical space become less robust as the scale of category diversity increases. Compared to GenPose, {\ModelName} achieves a significant lead by leveraging the powerful perception capabilities of the 2D foundation model, along with the robustness of clustering towards discrete symmetric properties. You can find qualitative visualizations in \Fref{fig:baseline_visualization}.

\vspace{-10pt}
\subsection{Ablation Study}

In order to validate the design decisions of our approach, we performed a series of ablation experiments on our method:

\begin{itemize}
    \item \textbf{w/o clustering.} Without clustering directly take the average of all remaining pose candidates after outlier removal as the pose estimation output.
    \item \textbf{w/o scale prediction.} Use the estimated pose to transform the observed point cloud into object space, then take the bounding box length as the maximum projection from the point cloud to each axis.
    \item \textbf{w/o simulated depth.}  Use perfect depth for training.
    \item \textbf{w/o point-wise feature fusion.} In the feature extraction stage, separately extract the RGB feature and geometric feature, and then concatenation.
\end{itemize}

Table \ref{table: ablation} illustrates the contribution of each component of {\ModelName} to its performance. 
The introduction of the clustering module allows GenPose++ to effectively aggregate the multimodal distributions caused by discrete symmetries, leading to higher performance. The scale prediction in GenPose++ significantly surpasses direct calculations from the object's point cloud due to ambiguities from partial observations and errors from point cloud noise, particularly in transparent and reflective objects. Training with simulated depth data results in better performance than training with perfect point clouds, as the physics-based depth camera simulation substantially reduces the sim2real gap for depth data. Point-wise fusion outperforms global fusion as it retains more of the object's local geometric features, which are crucial for accurate object pose prediction.

\begin{table}[t]
\centering
\caption{
\textbf{Ablation study on category-level 6D object pose estimation} 
}
\vspace{-5pt}
\input{Tables/cat_level_ablation}

\label{table: ablation}
\end{table}

\vspace{-10pt}
\subsection{Category-Level 6D Object Pose Tracking}

\subsubsection{Metric.}
We report the following metrics for object pose tracking evaluation:
\begin{itemize}
    \item \textbf{FPS}: Frames Per Second, which indicates the speed of pose tracking.
    \item \textbf{VUS@$5^\circ5$cm}: Volume Under Surface, assessing pose estimation accuracy for rotation errors within $0$ to $5^\circ$ and translation errors within $0$ to $5$cm.
    \item \textbf{mIoU}: Mean Intersection over Union, representing the average 3D overlap between the ground truth and the predicted bounding boxes.
    \item \textbf{Rerr($^\circ$)}: Average rotation error in degrees.
    \item \textbf{Terr(cm)}: Average translation error in centimeters.
\end{itemize}

\vspace{-10pt}
\subsubsection{Baselines.} This paper employs BundleTrack, CATRE, and GenPose as baselines for object pose tracking. BundleTrack is a training-free approach that utilizes multi-view feature point detection and matching for tracking the pose of unseen objects. CATRE aligns partially observed point clouds to abstract shape priors to estimate relative transformations, enhancing pose accuracy. Conversely, GenPose utilizes a generative approach to effectively resolve pose ambiguities, notably in symmetric objects. Following GenPose, we have adapted {\ModelName} to object pose tracking with minor modifications.

\vspace{-10pt}
\subsubsection{Results and Analysis.} ~\Tref{table: tracking} presents the results of category-level object pose tracking algorithms CATRE, GenPose, and our method, as well as the results for the unseen object pose tracking algorithm BundleTrack. The category-level pose estimation methods appear to achieve relatively better outcomes since they benefit from learning the category-level canonical space of objects within the SOPE dataset. However, for the training-free BundleTrack, reliance on RGB information for keypoint detection and matching poses challenges, often failing on objects with weak textures. Additionally, its dependency on depth values for global optimization renders it less effective in handling instances with substantial depth noise, such as transparent or specular objects. Our method, without undergoing specialized design, has achieved results comparable to state-of-the-art approaches. Although the inference speed of our method is lower than that of CATRE and GenPose, the achieved 17.8 FPS is sufficient for certain downstream tasks, such as robotic manipulation. Furthermore, recent rapid developments in research on fast samplers are beneficial to our approach, potentially enhancing its performance and applicability in real-time scenarios.

\begin{table}[h]
    \centering
    
\vspace{-10pt}
    \caption{\textbf{Results of category-level object pose tracking on {\RealData}.} The results are averaged over all 149 categories. \emph{GT. Pert.} denotes that a perturbed ground truth pose is utilized as the initial object pose.} 
    
\vspace{-5pt}
    \resizebox{\textwidth}{!}{
        \input{Tables/tracking_baselines}
    }
    \label{table: tracking}
    
\vspace{-10pt}
\end{table}

%% file: Tables/cat_level_baselines.tex

{
\newcommand{\A}{\multicolumn{2}{c|}{\multirow{2}{*}{Method}}}
\newcommand{\B}{\multirow{2}{*}{Input}}
\newcommand{\C}{\multirow{2}{*}{Prior-free}}
\newcommand{\K}{\multicolumn{3}{c|}{AUC $\uparrow$}}
\newcommand{\D}{\multicolumn{4}{c}{VUS $\uparrow$}}
\newcommand{\F}{\multicolumn{2}{c|}{}}
\newcommand{\G}{\multicolumn{1}{c|}{\multirow{4}{*}{Deterministic}}}
\newcommand{\I}{\multicolumn{1}{c|}{}}
\newcommand{\J}{\multicolumn{1}{c|}{\multirow{2}{*}{Probabilistic}}}
\newcommand{\Ours}{GenPose++(Ours)}
\newcommand{\colour}{\rowcolor{gray!15}}


\hskip-1.0cm\begin{tabular}{cc|cc|ccc|cccc}
\toprule
\A                   & \B     & \C         & \K                                      & \D                                                                 \\ \cmidrule{5-11}
\F                   &        &            & IoU$_{25}$  & IoU$_{50}$  & IoU$_{75}$  & $5^{\circ}2$cm & $5^{\circ}5$cm & $10^{\circ}2$cm & $10^{\circ}5$cm \\ \midrule \midrule
\G & NOCS~\cite{Wang_2019_CVPR}            & RGBD  & \Yes       & 0.0         & 0.0         & 0.0         & 0.0            & 0.0            & 0.0             & 0.0             \\
\I & SGPA~\cite{chen2021sgpa}            & RGBD  & \No        & 10.5        & 2.0         & 0.0         & 4.3            & 6.7            & 9.3             & 15.0            \\
\I & IST-Net~\cite{liu2023prior}         & RGBD  & \Yes       & 28.7        & 10.6        & 0.5         & 2.0            & 3.4            & 5.3             & 8.8             \\
\I & HS-Pose~\cite{Zheng_2023_CVPR}         & D     & \Yes       & 31.6        & 13.6        & 1.1         & 3.5            & 5.3            & 8.4             & 12.7            \\ \midrule
\J & GenPose~\cite{zhang2023genpose}         & D     & \Yes       & -           & -           & -           & 6.6            & 9.6            & 13.1            & 19.3            \\
\I & \Ours           & RGBD  & \Yes       & \textbf{39.0} & \textbf{19.1} & \textbf{2.0} & \textbf{10.0} & \textbf{15.1} & \textbf{19.5} & \textbf{29.4}  \\ \bottomrule
\end{tabular}
}

%% file: Tables/cat_level_ablation.tex
{
\newcommand{\A}{\multicolumn{2}{c|}{\multirow{2}{*}{Method}}}
\newcommand{\B}{\multirow{2}{*}{Data}}
\newcommand{\C}{\multirow{2}{*}{Prior-free}}
\newcommand{\D}{\multicolumn{7}{c}{VUS}}
\newcommand{\F}{\multicolumn{2}{c|}{}}
\newcommand{\G}{\multicolumn{1}{c|}{\multirow{4}{*}{Deterministic}}}
\newcommand{\I}{\multicolumn{1}{c|}{}}
\newcommand{\J}{\multicolumn{1}{c|}{\multirow{2}{*}{Probabilistic}}}
\newcommand{\Ours}{GenPose++(Ours)}
\newcommand{\colour}{\rowcolor{gray!15}}

\begin{tabular}{c|ccc|cccc}
\toprule
    Ablation                & \multicolumn{3}{c|}{AUC $\uparrow$}             & \multicolumn{4}{c}{VUS $\uparrow$}                      \\ \midrule
                            & IoU$_{25}$ & IoU$_{50}$ & IoU$_{75}$ & $5^{\circ}2$cm & $5^{\circ}5$cm & $10^{\circ}2$cm & $10^{\circ}5$cm \\ \midrule \midrule
Full                        &\textbf{39.0}&\textbf{19.1}&\textbf{2.0}&\textbf{10.0} &\textbf{15.1} &\textbf{19.5}  &\textbf{29.5}  \\ 
w/o scale prediction        & 13.8  & 3.4   & 0.2                 &\textbf{10.0} &\textbf{15.1} &\textbf{19.5}  &\textbf{29.5}  \\ 
w/o clustering              & 38.6  & 18.7  & 1.9                 &9.4  &14.1 &18.4  &27.8  \\ 
w/o simulated depth         & 35.3  & 16.5  & 1.7                 &7.3  &11.9 &14.6  &24.3  \\ 
w/o point-wise fusion       & 34.2  & 15.4  & 1.4                 &7.1  &10.6 &14.6  &21.9  \\ 
\bottomrule
\end{tabular}
}

%% file: Tables/tracking_baselines.tex
\begin{tabular}{c|*{6}{c|}c}
\toprule
Methods                                 & Input & Init.    & Speed(FPS)$\uparrow$ & $5^{\circ}5$cm$\uparrow$ & mIoU$\uparrow$ & $R_{err}$($^\circ$)$\downarrow$ & $T_{err}$(cm)$\downarrow$ \\ \midrule \midrule
BundleTrack~\cite{wen2021bundletrack}   & RGBD & GT. & 12.4 & 1.3 & 3.9 & 46.9 & 23.5\\
CATRE~\cite{liu2022catre}               & D    & GT. Pert.  & \textbf{38.5} & \textbf{15.9} & \textbf{55.4} & 21.3 & 2.6 \\
GenPose~\cite{zhang2023genpose}         & D    & GT. Pert.  & 26.3 & 13.3 &  -   & 19.3 & \textbf{1.2}   \\
GenPose++(Ours)                         & RGBD & GT. Pert.  & 17.8 & \textbf{15.9} &53.4  & \textbf{17.6}  & \textbf{1.2}\\
\bottomrule
\end{tabular}


%% file: Texs/6_conclusion.tex


In this study, we introduce {\Dataset}, a comprehensive dataset for 6D object pose estimation, featuring extensive scale, diversity, and material variety. Through thorough experimentation, our findings suggest that the probabilistic framework holds promise for category-level 6D object estimation, leveraging semantic information provided by RGB images to address large-scale pose estimation challenges.
However, the performance of {\ModelName} on {\Dataset} reveals significant room for improvement, with the model still hampered by slow inference speeds resulting from the iterative refinement nature inherent in the diffusion model. 
Future works could focus on addressing these challenges and integrating the universal 6D pose estimation module trained on {\Dataset} into a broader range of downstream tasks~\cite{wu2022targf, zeng2023distilling, cheng2023score, xue2023learning}.



%% file: SuppTexs/09_supp_Omni6DPose.tex
\begin{figure*}[tbp]
\begin{center}
\centerline{\includegraphics[width=\textwidth]{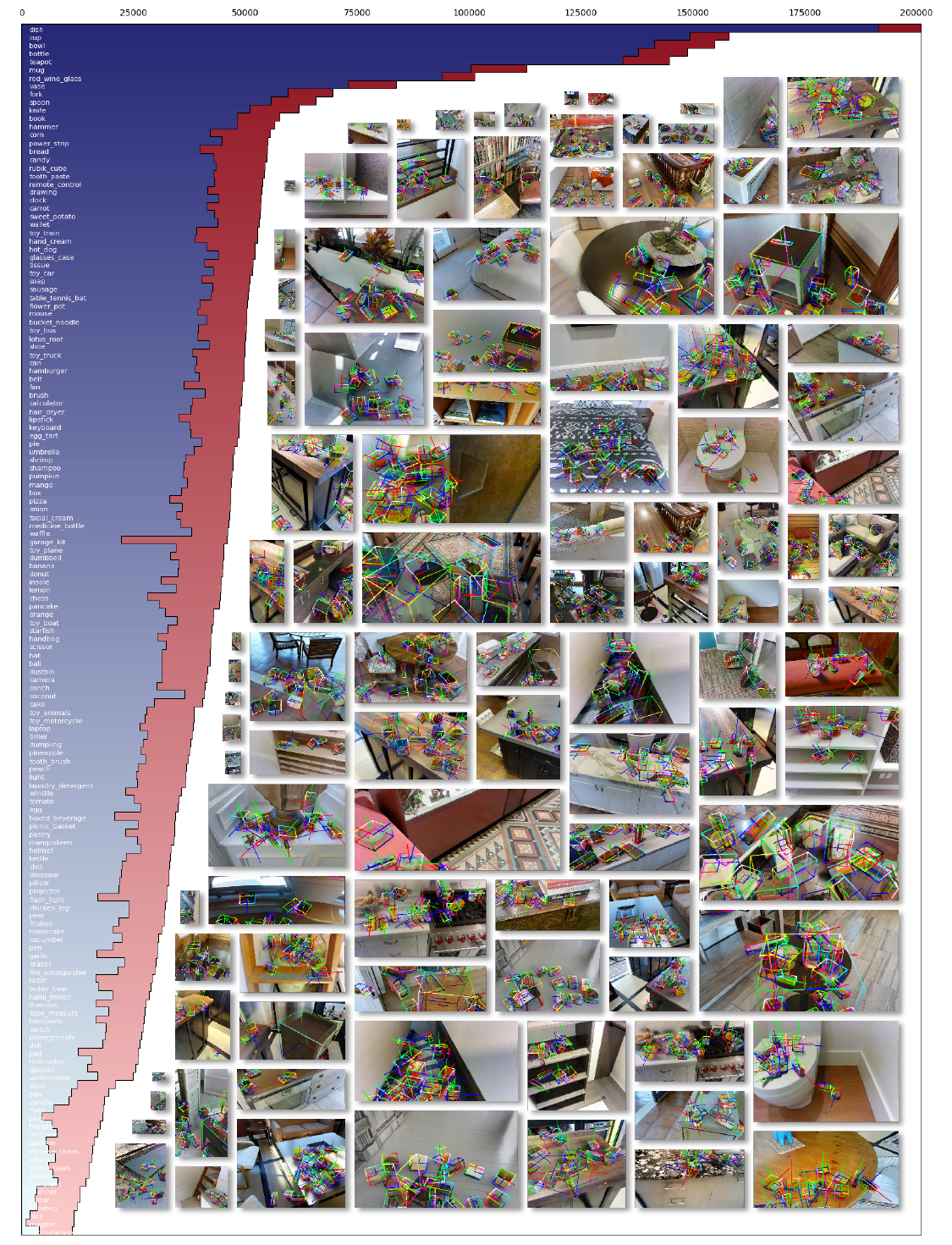}}
\caption{
Frequency of occurrence for all object categories within \textbf{\Dataset}.
}
\label{fig:full_categories}
\end{center}
\end{figure*}

\begin{figure*}[tbp]
\begin{center}
\centerline{\includegraphics[width=\textwidth]{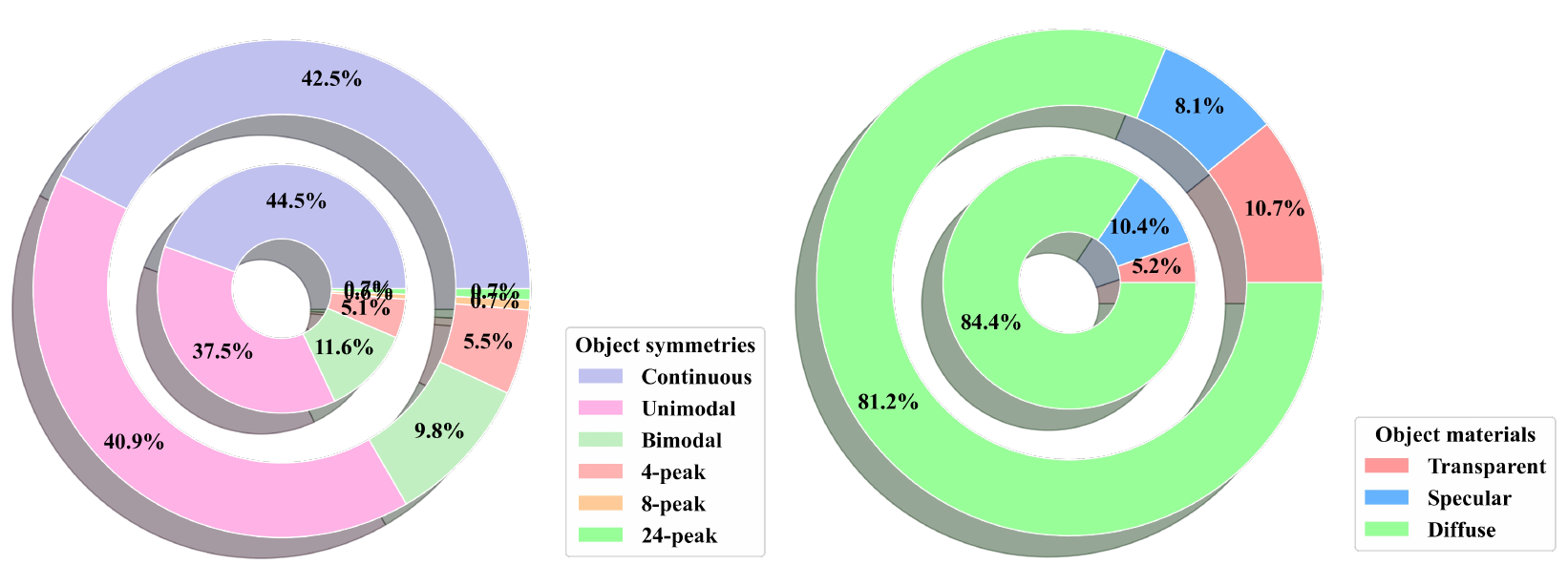}}
\caption{
Statistics of object symmetries and materials within \textbf{\Dataset}. Inner rings denote \textbf{\RealData} statistics and outer rings denote \textbf{\SimData}. The left chart categorizes object symmetries into 'continuous' for objects with continuous symmetry, 'unimodal' for objects with no symmetry attributes, and 'bimodal', '4-peak', '8-peak', and '24-peak' for objects with respective counts of discrete symmetry attributes. The right chart details object material distributions: transparent, specular, and diffuse.
}
\vspace{-15pt}
\label{fig:material_and_symmetry}
\end{center}
\end{figure*}

In this section, we provide additional statistical details about the Omni6DPose dataset, showcasing its characteristics and diversity. Specifically, it includes the statistics of full object categories, object symmetries and materials.

\vspace{-5pt}
\subsection{Statistics of Full Object Categories}
\Fref{fig:full_categories} illustrates the frequency of occurrence for all object categories within the Omni6DPose dataset, demonstrating the diversity of object categories covered. This diversity poses new challenges for universal 6D object pose estimation and is conducive to facilitating downstream applications, such as object rearrangement~\cite{wu2022targf, zeng2023distilling} and robot manipulation~\cite{an2023rgbmanip}. 

\vspace{-5pt}
\subsection{Statistics of Object symmetries}
In the domain of 6D object pose estimation, one of the principal challenges is mitigating the ambiguity issue arising from object symmetry. Omni6DPose includes a spectrum of objects characterized by distinct symmetry attributes, broadly classified into three categories: asymmetric objects such as cameras, continuously symmetric objects exemplified by bottles, and discretely symmetric objects, typical examples being boxes. Further delineation within Omni6DPose segregates discretely symmetric objects based on the count of peaks in the distribution of the objects' poses, categorized into Bimodal, 4-peak, 8-peak, and 24-peak classifications. The detailed statistical outcomes are illustrated in the left section of \Fref{fig:material_and_symmetry}. This vast diversity of object symmetries compels the development of new strategies and techniques for precise 6D object pose estimation.

\vspace{-5pt}
\subsection{Statistics of Object Materials}
Objects in daily life are made from diverse materials, such as transparent glass mugs and reflective knives. Precise 6D object pose estimation across different materials is crucial for the application of pose estimation in real-world scenarios. Omni6DPose, serving as a comprehensive large-scale 6D object pose dataset, includes a diverse range of materials, categorized into three main types: Diffuse objects, Transparent objects, and Specular objects. The distribution of each material type within the ROPE and SOPE subsets of the dataset is detailed in the right of \Fref{fig:material_and_symmetry}. This variety provides a significant dataset for research into 6D pose estimation of objects with challenging material properties.


%% file: SuppTexs/11_supp_Genpose++_details.tex
\subsection{Training Details}
In the training phases, both the ScoreNet and Energy models are subjected to training with a batch size of $128$, employing the Adam optimizer. The initial learning rate is established at $10^{-3}$, subsequently decaying to $10^{-4}$ to foster optimal convergence. Specifically, ScoreNet is trained for a total of $28$ epochs, whereas the Energy model undergoes $25$ epochs of training. 

\subsection{Network Details}
\vspace{-5pt}
In this section, we detail the feature encoder of GenPose++, which processes data from two modalities: RGB and pointcloud. 
For the RGB modality, we utilize a frozen, pre-trained DINOv2~\cite{oquab2023dinov2} to extract the semantic features. Specifically, we begin by cropping the object region from the original image based on the object mask and resizing this crop to $224 \times 224$ pixels. This resized region is then passed through DINOv2 to produce a feature map of dimensions $16 \times 16$. Each feature vector in this map is $384$ elements long and represents a $14 \times 14$ patch from the original RGB image. To streamline the process, we employ the `ViT-S/14' variant of DINOv2, which reduces the number of parameters and enhances inference speed.
For the pointcloud modality, the object's point cloud is extracted directly using the object mask. We then apply Farthest Point Sampling (FPS) to sample $1024$ points and extract global features using pointnet++~\cite{zhang2023genpose}. During the feature extraction process for the point cloud, the RGB features are point-wise concatenated onto the corresponding points, integrating data from both modalities to enrich the feature representation.

%% file: SuppTexs/10_supp_SOPE.tex

\begin{figure*}[tbp]
\begin{center}
\centerline{\includegraphics[width=\textwidth]{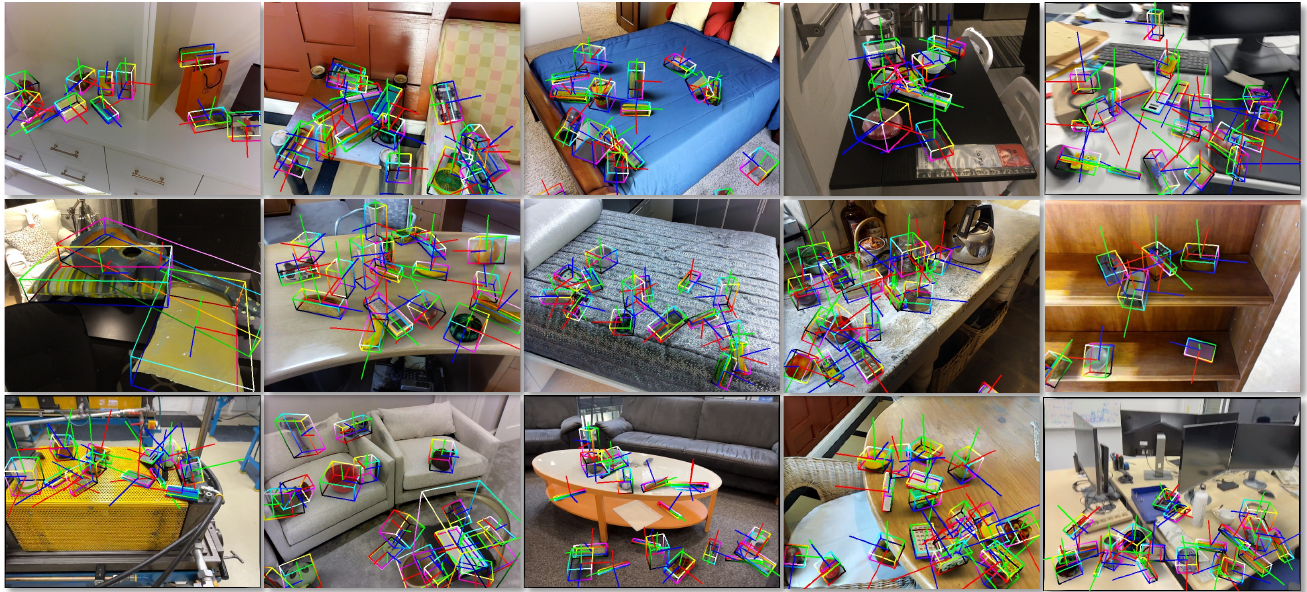}}
\vspace{-5pt}
\caption{
\textbf{{\SimData} dataset visualization.} In the figure, bounding boxes are colored according to the coordinates in the object's coordinate system.
}
\vspace{-30pt}
\label{fig:SOPE_visualization}
\end{center}
\end{figure*}

\vspace{-5pt}
We synthesize {\SImgNum} frames for training by integrating context-aware mixed reality with physics-based depth sensor simulation. To enhance the generalization capability of {\SimData}, we systematically apply domain randomization during the data generation process, specifically targeting variations in illumination and object material properties. Considering the relatively lower instance numbers of transparent and reflective objects among all types of objects, we increase their occurrence probability in {\SimData}. Consequently, \Fref{fig:SOPE_visualization} exhibits selected examples from the Synthetic Objects in {\SimData}, showcasing the diversity and realism of the simulated dataset.

%% file: SuppTexs/12_supp_add_experiments.tex
In this section, we analyze the necessity of physics-based deep simulation and the distance in feature space between context-aware mixed reality generated RGB images and real images. This elucidates why the SOPE dataset enhances sim-to-real generalization capabilities.

\subsection{Physical-based Depth Sensor Simulation.}
\begin{figure*}[tbp]
\begin{center}
\centerline{\includegraphics[width=\textwidth]{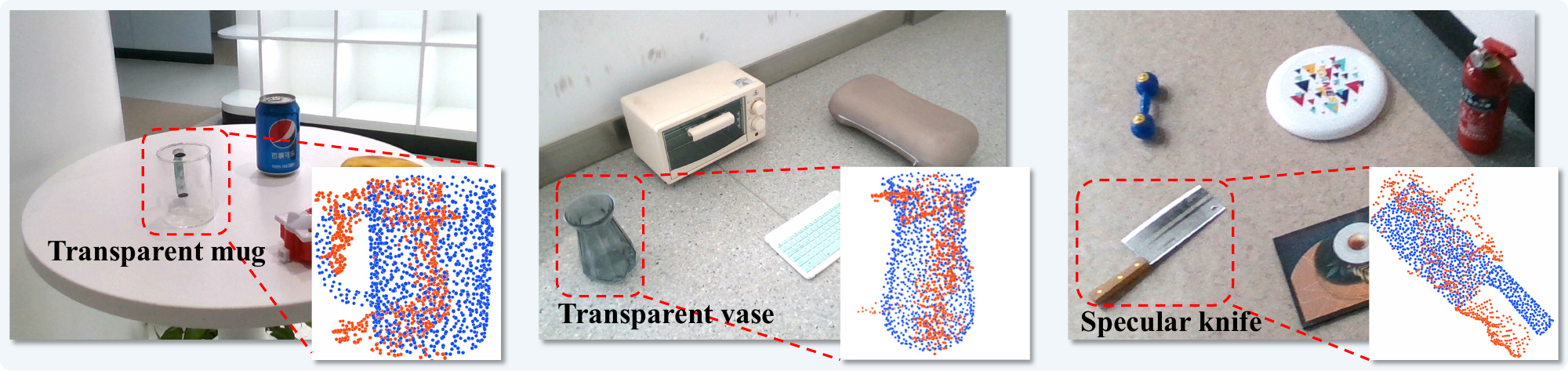}}
\vspace{-5pt}
\caption{
Visualization of structured-light depth sensor noise on transparent and specular areas. The visualization presents the discrepancy between the ground truth pointcloud (blue) and the captured pointcloud (red) by the depth sensor. The examples include a transparent mug, a transparent vase, and a specular knife.
}
\vspace{-30pt}
\label{fig:depth_noise}
\end{center}
\end{figure*}

Structured light-based depth sensors typically introduce noise into the captured depth images, which is particularly pronounced in regions with transparent and reflective objects. This results in a considerable sim-to-real gap when training on perfect synthetic point clouds. Our ablation experiments, as discussed in the main manuscripts, have already established that physics-based depth sensor simulations can significantly bridge the sim-to-real gap. To more vividly demonstrate the divergence between the point clouds captured by the depth sensor and the ideal synthetic ones, \Fref{fig:depth_noise} shows the depth noise in the transparent and reflective regions from a subset of the ROPE dataset. These visualizations clearly articulate the necessity of physics-based depth sensor simulation.

\subsection{Context-Aware Mixed Reality RGB.}



Previous synthetic datasets employ rasterization to integrate manually-created object models into a real scene, an approach that falls short in terms of overall image fidelity and the realism of individual objects.
In contrast, our method 
\begin{wrapfigure}{hr}{0.65\textwidth}
\centering
\includegraphics[width=0.65\textwidth]{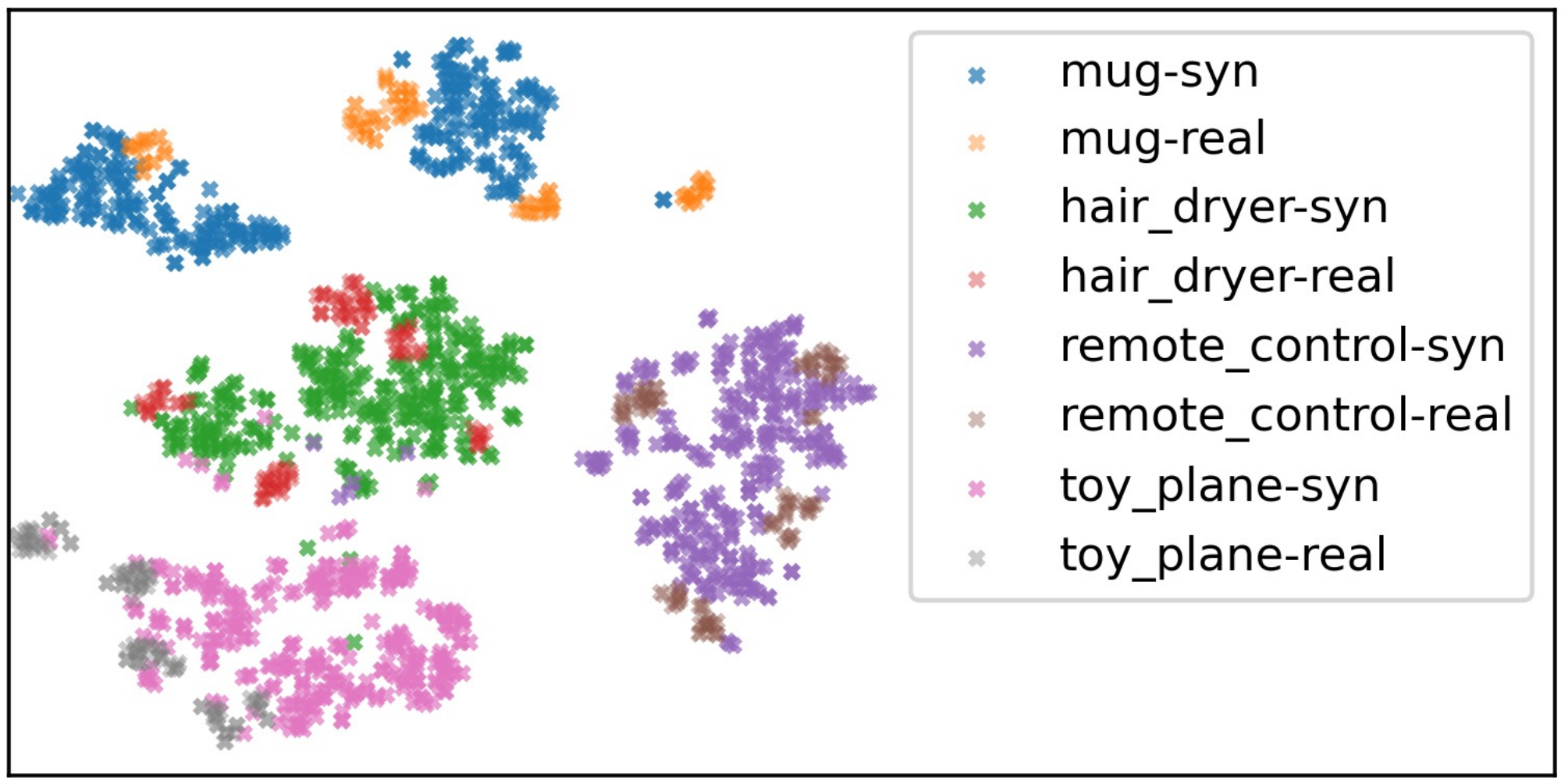}
\vspace{-4pt}
\caption{Visualization of the features of RGB images extracted by DINOv2, reduced to 2D plane using t-SNE.}
\vspace{-20pt}
\label{fig:rgb_sim-to-real}
\end{wrapfigure}
leverages ray-tracing and real scanned objects to produce highly realistic imagery. As noted in the main manuscript, the inclusion of RGB information markedly enhances performance. To delve deeper into this, in \Fref{fig:rgb_sim-to-real}, we showcase the comparison of features extracted using DINOv2 from both synthetic and real RGB images. It demonstrates that the features within the synthetic data set significantly overlap with those in the real data, which bridges the semantic sim-to-real gap.
